\documentclass[10pt,twocolumn,letterpaper]{article}

\usepackage[pagenumbers]{cvpr} 

\usepackage{graphicx}
\usepackage{amsmath}
\usepackage{amssymb}
\usepackage{booktabs}
\usepackage{algorithm,algpseudocode}
\usepackage{listings}
\usepackage{multirow}
\usepackage{bbm}
\usepackage{verbatim}
 
\usepackage[pagebackref,breaklinks,colorlinks]{hyperref}

\usepackage[capitalize]{cleveref}
\crefname{section}{Sec.}{Secs.}
\Crefname{section}{Section}{Sections}
\Crefname{table}{Table}{Tables}
\crefname{table}{Tab.}{Tabs.}

\def\cvprPaperID{15} 
\def\confName{CVPR}
\def\confYear{2022}

\begin{document}
	
\title{Benchmarking Unsupervised Anomaly Detection and Localization}

\author{
    Ye Zheng$^{2}$ \quad
    Xiang Wang$^{1}$ \quad
    Yu Qi$^3$ \quad
    Wei Li$^1$ \quad \\
    Liwei Wu$^1$ \quad\\[8pt]
    $^2$University of Chinese Academy of Sciences \quad
    $^1$SenseTime Research \quad
    $^3$Tsinghua University \quad \\[5pt]
    }

\maketitle

\begin{abstract}
Unsupervised anomaly detection and localization, as of one the most practical and challenging problems in computer vision, has received great attention in recent years. From the time the MVTec AD dataset was proposed to the present, new research methods that are constantly being proposed push its precision to saturation. It is the time to conduct a comprehensive comparison of existing methods to inspire further research. This paper extensively compares 13 papers in terms of the performance in unsupervised anomaly detection and localization tasks, and adds a comparison of inference efficiency previously ignored by the community. Meanwhile, analysis of the MVTec AD dataset are also given, especially the label ambiguity that affects the model fails to achieve full marks. Moreover, considering the proposal of the new MVTec 3D-AD dataset, this paper also conducts experiments using the existing state-of-the-art 2D methods on this new dataset, and reports the corresponding results with analysis.
\end{abstract}


\section{Introduction}
\label{sec:intro}
\begin{figure}[tp]
    \centering
    \includegraphics[width=1.0\linewidth]{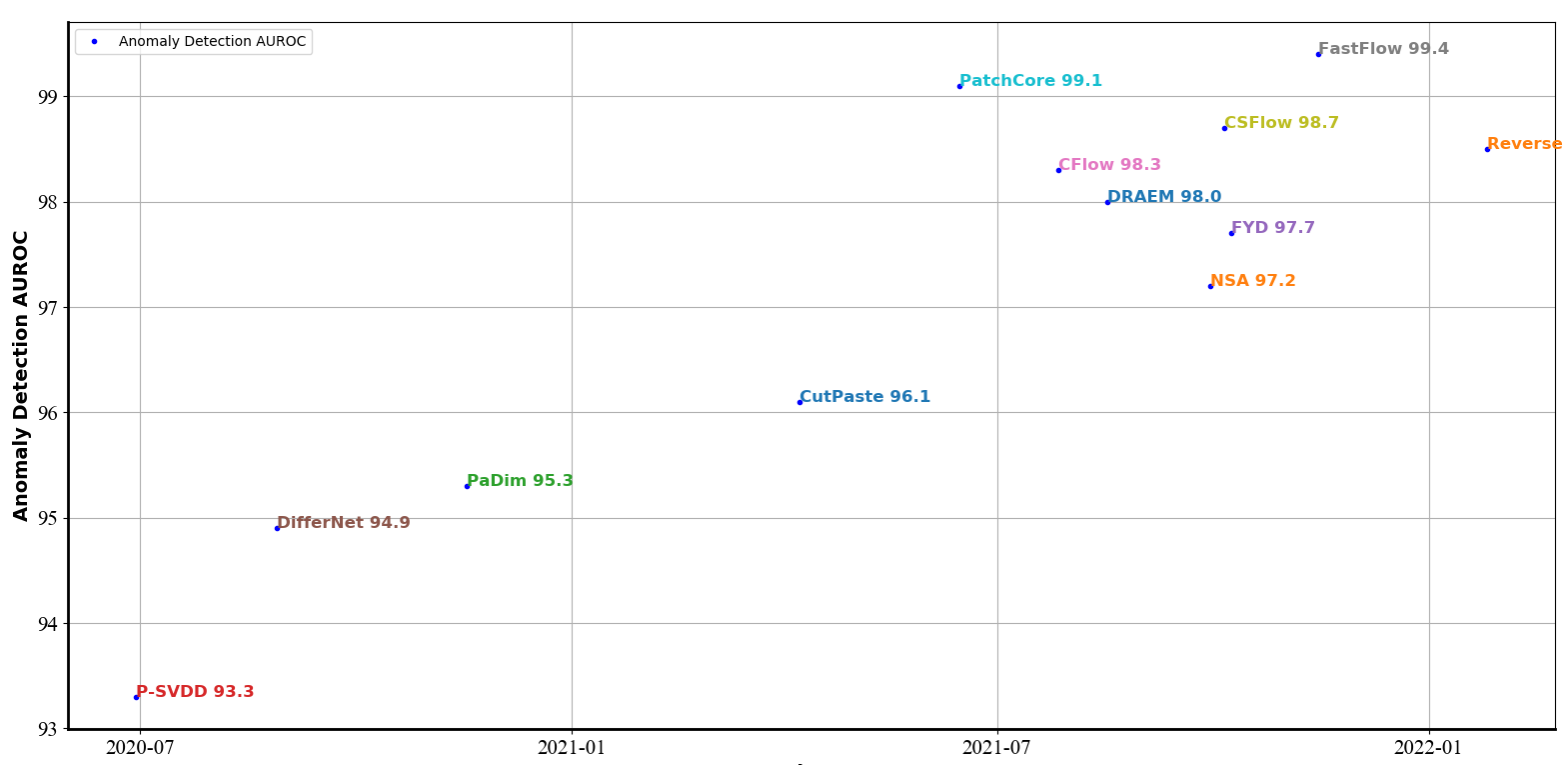}
    \caption{Performance of anomaly detection on MVTec AD dataset versus time curve.}
    \label{fig:1}
\end{figure}

Image anomaly detection and localization is an important computer vision task that deals with detecting and localization unexpected or abnormal patterns in digital images, which has been widely used in manufacturing~\cite{bergmann2019mvtec} and healthcare~\cite{seebock2016identifying}. Limited by the data scarcity problem of anomaly data and the 
extremely uneven distribution among normal and abnormal classes, supervised learning based methods are difficult to apply, especially deep learning methods. To this end, unsupervised anomaly detection and localization task has been proposed, which aims to develop computational models and techniques that detect and localize anomalies without using any anomaly data during training. Previous studies focus on the unsupervised anomaly detection task and success in abstracting semantically rich representation for isolating defect images, nonetheless, they lack the ability to explore the fine-grained anomalies. For example, previous works~\cite{golan2018deep,sohn2020learning} usually set one category in CIFAR-10 dataset~\cite{krizhevsky2009learning} as the normal class and the rest as anomalies. However, in the real world of manufacturing or medicine, the distinction between normal and abnormal images is finer and subtler than these object class distinctions~\cite{bergmann2019mvtec}. Driven by this problem, more practical benchmarks has been proposed, the most influential one is the MVTec AD dataset~\cite{bergmann2019mvtec}, which provides several fine-grained anomalies with corresponding pixel-level anomaly segmentation annotation. After that, a large number of works have been proposed for improving the performance in anomaly detection and localization task. As shown in Fig.~\ref{fig:1}, up to now, state-of-the-art methods achieve 99.4\% AUC in image-level anomaly detection and 98.6\% in pixel-level anomaly localization, and make the performance of the benchmark close to saturation. Contrary to the near-saturation in performance, there has not been a comprehensive comparison of existing methods, therefore, we extensively compare 13 papers in terms of the performance in unsupervised anomaly detection and localization tasks. Moreover, we add the comparison of inference efficiency previously ignored by the community. A comprehensive comparison of existing methods will help subsequent researchers to make more reasonable comparisons and promote the development of the community.

Recently, a new dataset MVTec 3D-AD dataset~\cite{bergmann2021mvtec} has been proposed to fill the gap that lack of unsupervised anomaly detection and localization benchmark in 3D space and to spark further interest in the development of new methods. This dataset contains a set of exclusively anomaly-free 3D scans of an object and various anomaly data in 3D space. We conduct experiments using the existing 2D state-of-the-art methods on this new dataset to explore the effectiveness of existing state-of-the-art 2D methods on the new 3D dataset when they only use RGB data like in 2D task. We find that some existing 2D methods can also achieve a high performance on 3D dataset.

In summary, our main contributions are: 
\begin{itemize}
	\item We present a comprehensive comparison of existing methods for unsupervised anomaly detection and localization task. In addition to performance comparisons, we also present previously overlooked inference efficiency. We believe this will help other researchers to quickly obtain comparative results with existing methods and facilitate subsequent studies.
	\item We summarize the examples of label ambiguity in the current MVTec AD dataset that affect the model evaluation results. We find that after correcting these annotations, the performance of the state-of-the-art method is basically close to a full score.
	\item We conduct experiments to evaluate existing state-of-the-art 2D methods in the new MVTec 3D-AD dataset, and we find that some existing 2D methods can also achieve satisfactory results.
\end{itemize}

\section{Related Work}\label{sec:related}


\begin{figure}
    \centering
    \includegraphics[width=0.5\linewidth]{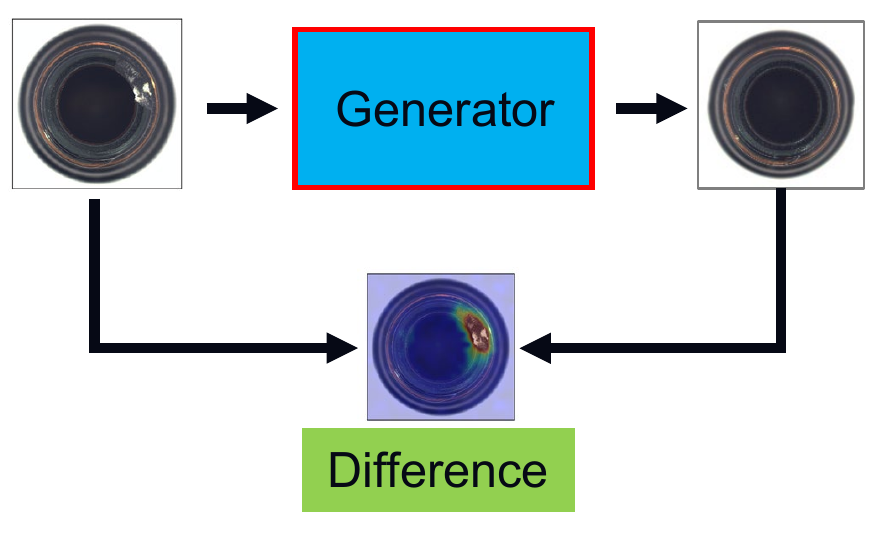}
    \caption{The methodology pipeline of reconstruction-based unsupervised anomaly detection and localization methods.}
    \label{fig:2}
\end{figure}

\begin{figure}
    \centering
    \includegraphics[width=0.95\linewidth]{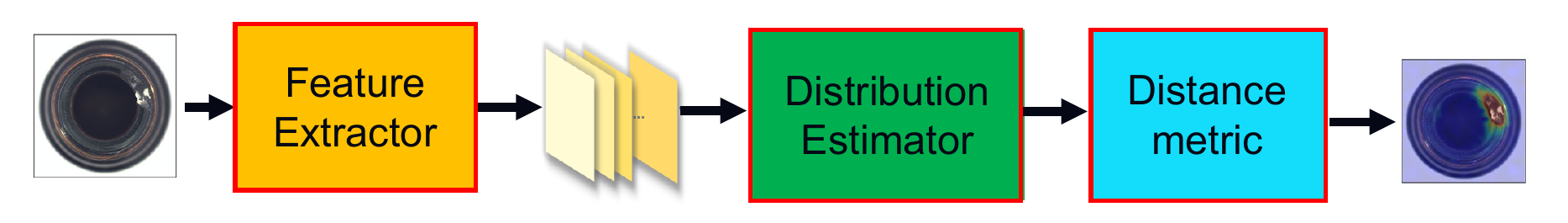}
    \caption{The methodology pipeline of representation-based unsupervised anomaly detection and localization methods.}
    \label{fig:3}
\end{figure}

The mainstream unsupervised anomaly detection methods are either reconstruction-based or representation-based, the methodology pipeline of these two schools is shown in Fig.~\ref{fig:2} and Fig.~\ref{fig:3}, respectively. 

\subsection{Reconstruction-based methods}
Reconstruction-based methods usually adopt an image generator such as auto-encoders~\cite{bergmann2019mvtec, gong2019memorizing}, generative adversarial networks~\cite{schlegl2017unsupervised, sabokrou2018adversarially, zenati2018efficient}, the combination of both auto-encoders and GANs~\cite{akccay2019skip, pidhorskyi2018generative}, and the student-teacher model~\cite{bergmann2020uninformed, wang2021glancing}
to generate or reconstruct the input image and localize the abnormal area through the difference between input and reconstructed images. DD As shown in Fig.~\ref{fig:2}, these methods focus on modeling the manifold of normal images and reconstruction from the latent subspace. They hold the insights that the anomalies can not be reconstructed since they are not observed in training. In inference, anomalous regions are spotted when the reconstructed image diverges from the original one. The pixel-wise reconstruction error can be applied to localize anomalies~\cite{bergmann2019mvtec}, and the image level anomaly score is thus determined by aggregating pixel-wise errors~\cite{gong2019memorizing} or thresholding the discriminator output~\cite{akcay2018ganomaly}. One of the most recent works, DRAEM~\cite{zavrtanik2021draem} applies denoising auto-encoders (DAE) to corrupt a normal image and learn to reconstruct the original, which also achieves a satisfactory performance. 

\subsection{Representation-based methods}
Representation-based methods adopt a different methodology in Fig.~\ref{fig:3}, which first uses different \textbf{feature extractors} to obtain discriminative features for normal images~\cite{ruff2018deep,bergman2020classification,rippel2021modeling,rudolph2021same} or normal image patches~\cite{yi2020patch,cohen2020sub,reiss2021panda,gudovskiy2021cflow}, and then establish the distribution of these normal features. Finally, the anomaly score is calculated by the distance between the embedding of a test image and the distribution of normal image representations. Therefore, the \textbf{feature extractor}, the \textbf{distribution estimator}, and the \textbf{distance metric} are three main components in representation-based models. 

For the \textbf{feature extractor}, existing approaches usually adopt the CNN based or ViT based model~\cite{defard2021Padim,cohen2020sub,roth2021towards,yu2021fastflow}, such as the ResNet~\cite{he2016deep} or ViT~\cite{dosovitskiy2020image} trained on the ImageNet. Furthermore, to learn the domain-specific semantic vectors for images, a series of methods~\cite{li2021cutpaste,zheng2021focus,reiss2021panda} employed self-supervised learning to achieve the ImageNet pretrained model adaption. These methods train the feature extractor by proposing different pretext tasks. CutPaste~\cite{li2021cutpaste} and NSA~\cite{schluter2021self} cut a patch from one image and paste it on the other to construct the negative sample and uses a normal-abnormal binary classification task to train the model. However, the representation of created negative irregularities usually does not overlap with real-world anomalies~\cite{li2021cutpaste}, which limits the generalization potential of these methods in inference processes. FYD~\cite{zheng2021focus} takes into account spatial consistency on image representations to detect and localize the fine-grained defects. It proposes a non-contrastive learning method to train the model to avoid constructing negative samples and learn a dense and compact distribution from normal images with a coarse-to-fine alignment process. PANDA~\cite{reiss2021panda} uses simple early stopping, sample-wise early stopping, and continual learning to avoid the model collapse in training.

For the \textbf{distribution estimator}, previous methods usually use non-parametric methods and the distribution of normal image representation is typically characterized by the center of an n-sphere for the normal image~\cite{ruff2018deep}, the Gaussian distribution of normal image, or the kNN for the entire normal image embedding~\cite{cohen2020sub,defard2021Padim}. PaDim~\cite{defard2021Padim} models the normal features as a multidimensional Gaussian distribution and uses the Mahalanobis distance to measure the distance between the test sample and the distribution. PatchCore~\cite{roth2021towards} uses a maximally representative memory bank of normal patch-features and uses the k-nearest neighbor method to search for the closest feature for distance metric.
Recently, some works~\cite{rudolph2021same,gudovskiy2021cflow,yu2021fastflow} began to use parametric methods such as normalizing flow~\cite{dinh2014nice,kingma2018glow} to estimate the distribution of normal images. Through a trainable process that maximizes the log-likelihood of normal image features, they embed normal image features into standard normal distribution and use the probability to identify and locate anomalies. DifferNet~\cite{rudolph2021same} achieved good anomaly detection performance in image level by using NF to estimate the precise likelihood of test images. Unfortunately, this work failed to obtain the exact anomaly localization results since global average pooling was used. CFLOW-AD~\cite{gudovskiy2021cflow} proposes to use hard code position embedding to leverage the distribution learned by NF, which probably underperforms at more complicated datasets. CSFlow~\cite{rudolph2022fully} proposes a novel fully convolutional cross-scale normalizing flow to jointly processes multiple feature maps of different scales. FastFlow~\cite{yu2021fastflow} designs a 2D normalizing flow for anomaly detection and localization with fully convolutional networks and achieves the state of the art performance.

\section{Experiments}
In this section, we demonstrate our benchmarking procedures to present the comprehensive comparison of existing methods.
Firstly, we introduce 2D and 3D datasets for unsupervised anomaly detection and localization scenario and their corresponding evaluation criterion.
%
Then we give the performance comparison of recent mainstream unsupervised anomaly detection methods in 2D dataset, including both image-level and pixel-level results. In addition, we also report the inference efficiency of several state-of-the-art methods. Moreover, the experimental results of 3D dataset are also provided.

\subsection{Datasets}

\subsubsection{MVTec AD 2D} 
The MVTec Anomaly Detection dataset~\cite{bergman2020classification} comprises 15 categories with 3629 images for training and 1725 images for testing. This dataset contains 15 categories, in which 10 of them are objects and the others are textures. The training set contains only images without defects, while the test set contains both normal images and abnormal images.

Under the unsupervised setting, the model for each category is trained with its respective normal samples and evaluated in test set images which contain both normal and abnormal samples.

\subsubsection{MVTec AD 3D} 
We also perform extensive experiments on MVTec 3D Anomaly Detection~\cite{bergmann2021mvtec} which is recently released. 
As far as we know, this dataset is the only publicly available industrial dataset in community, which is designed to be a benchmark for unsupervised anomaly detection in 3D point cloud spaces.
It contains over 4000 high-resolution 3D scans of 10 categories of industrially manufactured products, among which five of them expose natural changes(bagel, carrot, cookie, peach, and potato). Two of the classes are rigid (cable gland, dowel), and the others (foam, rope, tire) are manufactured products with structural or texture anomalies.
The anomalies can be divided into five types (hole, contaminated, crack, cut, combined) roughly, some classes have unique anomalies such as color and open.

Under the same setting as 2D task, we need to train the model for each category with its respective normal samples and evaluate it in test set images which contain both normal and abnormal samples.

\subsection{Benchmarking details}

%

%
\textbf{For the 2D unsupervised anomaly detection and localization, } we curate and present the performance of recent state-of-the-art approaches on the MVTec AD dataset, including reconstruction-based and representation-based methods. Considering the inference efficiency is vital for practical applications, we construct experiments to compare the inference speed of several state-of-the-art methods. For fair comparison, we implemented a unified data interface for unsupervised anomaly detection and localization task, which consists of a unified data loading utility and a unified inference paradigm. The hardware configuration
of the machine used for testing is Intel(R) Xeon(R)
CPU E5-2680 V4@2.4GHZ and NVIDIA GeForce GTX
1080Ti.

\textbf{For the 3D unsupervised anomaly detection and localization,} we implement and evaluate the state-of-the-art approaches on the new 3D dataset. In order to explore the effect of these 2D methods in 3D data when still using RGB information for abnormal detection and localization, we use the RGB data only in this dataset to train and evaluate the model.
%



\begin{table*}[tbp]
\setlength\tabcolsep{2pt}
\resizebox{\linewidth}{!}{
\begin{tabular}{l|cc|ccccccccccc}
\toprule
\multicolumn{1}{c|}{\multirow{2}{*}{Category}} & \multicolumn{2}{l|}{Reconstruction based} & \multicolumn{11}{c}{Representation based} \\ \cline{2-14} 
\multicolumn{1}{c|}{} & DRAEM & R-D & PaDim & P-SVDD & FYD & SPADE & CutPaste & NSA & DifferNet & CFlow & FastFlow & CSFlow & PatchCore \\ \hline
carpet & 97.0 & 98.9 & - & 92.9 & 98.8 & - & 93.9 & 95.6 & 84.0 & \textbf{100.0} & \textbf{100.0} & \textbf{100.0} & \textbf{100.0} \\
grid & 99.9 & \textbf{100.0} & - & 94.6 & 98.9 & - & \textbf{100.0} & 99.9 & 97.1 & 97.6 & 99.7 & 99.0 & 98.2 \\
leather & \textbf{100.0} & \textbf{100.0} & - & 90.9 & \textbf{100.0} & - & 100.0 & 99.9 & 99.4 & 97.7 & \textbf{100.0} & \textbf{100.0} & \textbf{100.0} \\
tile & 99.6 & 99.3 & - & 97.8 & 98.8 & - & 94.6 & \textbf{100.0} & 92.9 & 98.7 & \textbf{100.0} & \textbf{100.0} & 98.7 \\
wood & 99.1 & 99.2 & - & 96.5 & 99.4 & - & 99.1 & 97.5 & 99.8 & 99.6 & \textbf{100.0} & \textbf{100.0} & 99.2 \\ \hline
Average & 99.1 & 99.5 & 98.8 & 94.5 & 99.2 & - & 97.5 & 98.6 & 94.6 & 98.7 & 99.9 & 99.8 & 99.2 \\ \hline
bottle & 99.2 & \textbf{100.0} & - & 98.6 & \textbf{100.0} & - & 98.2 & 97.7 & 99.0 & \textbf{100.0} & \textbf{100.0} & 99.8 & \textbf{100.0} \\
cable & 91.8 & 95.0 & - & 90.3 & 95.3 & - & 81.2 & 94.5 & 86.9 & \textbf{100.0} & \textbf{100.0} & 99.1 & 99.5 \\
capsule & 98.5 & 96.3 & - & 95.8 & 92.5 & - & 98.2 & 95.2 & 88.8 & 99.3 & \textbf{100.0} & 97.1 & 98.1 \\
hazelnut & \textbf{100.0} & 99.9 & - & 92.0 & 99.9 & - & 98.3 & 94.7 & 99.1 & 96.8 & \textbf{100.0} & 99.6 & \textbf{100.0} \\
metal nut & 98.7 & \textbf{100.0} & - & 94.0 & 99.9 & - & 99.9 & 98.7 & 95.1 & 91.9 & \textbf{100.0} & 99.1 & \textbf{100.0} \\
pill & 98.9 & 96.6 & - & 86.1 & 94.5 & - & 94.9 & 99.2 & 95.9 & \textbf{99.9} & 99.4 & 98.6 & 96.6 \\
screw & 93.9 & 97.0 & - & 81.3 & 90.1 & - & 88.7 & 90.2 & \textbf{99.3} & 99.7 & 97.8 & 97.6 & 98.1 \\
toothbrush & \textbf{100.0} & 99.5 & - & \textbf{100.0} & \textbf{100.0} & - & 99.4 & \textbf{100.0} & 96.1 & 95.2 & 94.4 & 91.9 & \textbf{100.0} \\
transistor & 93.1 & 96.7 & - & 91.5 & 99.2 & - & 96.1 & 95.1 & 96.3 & 99.1 & 99.8 & 99.3 & \textbf{100.0} \\
zipper & \textbf{100.0} & 98.5 & - & 97.9 & 97.5 & - & 99.9 & 99.8 & 98.6 & 98.5 & 99.5 & 99.7 & 99.4 \\ \hline
Average & 97.4 & 98.0 & 93.6 & 92.8 & 96.9 & - & 95.5 & 96.5 & 95.5 & 98.0 & 99.1 & 98.2 & \textbf{99.2} \\ \hline
Average-all & 98.0 & 98.5 & 95.3 & 93.3 & 97.7 & 85.5 & 96.1 & 97.2 & 94.9 & 98.3 & \textbf{99.4} & 98.7 & 99.1 \\ \bottomrule
\end{tabular}
}
\caption{Anomaly detection performance (Image-level AUCROC) on MVTec AD dataset~\cite{bergmann2019mvtec}.}
\label{table:imgroc}
\end{table*}

\subsubsection{Evaluation Metrics }
For anomaly detection at image level, we use ROCAUC in this paper.
To assess the anomaly localization performance of a method on MVTec AD 3D~\cite{bergmann2021mvtec}, we use the PRO~\cite{bergmann2021mvtec} to define the overlap between the binary predictions and the ground truth. 
We firstly require the methods to output a  anomaly score for each pixel of the test set. These anomaly scores are converted to binary predictions using a threshold.
The PRO value is computed as

\begin{equation}
    \mathrm{PRO} = \frac{1}{K} \sum_{k=1}^{K} \frac{\left|P \cap C_{k}\right|}{\left|C_{k}\right|}
    \label{method:pro}
\end{equation}
where K is the total number of ground truth components. 
Given different threshold, the curve is produced by the resulting PRO values and its corresponding false positive rates.
The final performance measure is computed by integrating this curve up to a limited false positive rate and normalizing the resulting area to the interval [0,1].  
When small anomalies dominate in the dataset, the PRO is particularly useful in evaluation.

\section{Results and Discussion }
\label{sec:Main Results}
For MVTec AD dataset for 2D task, we report the performance comparison of Glance~\cite{wang2021glancing}, DRAEM~\cite{zavrtanik2021draem}, DFR~\cite{yang2020dfr}, R-D~\cite{deng2022anomaly}, PaDim~\cite{defard2021Padim}, P-SVDD~\cite{yi2020patch}, FYD~\cite{zheng2021focus}, SPADE~\cite{cohen2020sub}, PANDA~\cite{reiss2021panda}, CutPaste~\cite{li2021cutpaste}, NSA~\cite{schluter2021self}, CFlow~\cite{gudovskiy2021cflow}, FastFlow~\cite{yu2021fastflow}, PatchCore~\cite{roth2021towards} in terms of the image-level and pixel-level metrics. The inference efficiencies of some of these methods are also provided. For MVTec 3D-AD dataset for 3D task, we give the experiment results of DifferNet~\cite{rudolph2021same}, PaDim~\cite{defard2021Padim}, PatchCore, CFlow~\cite{gudovskiy2021cflow}, CSFlow~\cite{rudolph2022fully} and FastFlow~\cite{yu2021fastflow}.

\subsection{MVTec AD 2D}
\subsubsection{Performance Comparison Results}
The anomaly detection performance comparison results of mainstream 2D methods are shown in Table~\ref{table:imgroc}, which are evaluated with the image-level AUROC metric, and Table~\ref{table:pixroc} reports the pixel-level results. We can observe that:
\begin{itemize}
    \item The representation-based methods have better performance than the reconstruction-based methods in general.
    \item Modelling the normal distribution through the feature embedding memory bank or the flow model achieve the best performance.
    \item Several state-of-the-art methods approach full performance.
\end{itemize}

\begin{table*}[htbp]
\setlength\tabcolsep{2pt}
\resizebox{\linewidth}{!}{

\begin{tabular}{l|cccc|cccccccccc}
\toprule

\multicolumn{1}{c|}{\multirow{2}{*}{Catagory}} & \multicolumn{4}{c|}{Reconstruction based} & \multicolumn{10}{c}{Representation based} \\ \cline{2-15} 
\multicolumn{1}{c|}{} & \multicolumn{1}{c}{Glance} & \multicolumn{1}{c}{DRAEM} & \multicolumn{1}{c}{DFR} & \multicolumn{1}{c|}{R-D} & \multicolumn{1}{c}{Padim} & \multicolumn{1}{c}{P-SVDD} & \multicolumn{1}{c}{FYD} & \multicolumn{1}{c}{SPADE} & \multicolumn{1}{c}{PANDA} & \multicolumn{1}{c}{CutPaste} & \multicolumn{1}{c}{NSA} & \multicolumn{1}{c}{CFlowAD} & \multicolumn{1}{c}{Fastflow} & \multicolumn{1}{c}{Patchcore} \\ \hline
carpet & 96.0 & 95.5 & 97.0 & 98.9 & 99.1 & 92.6 & 98.5 & 97.5 & 98.6 & 98.3 & \textbf{95.5} & 99.3 & 99.4 & 99.0 \\
grid & 78.0 & 99.7 & 98.0 & \textbf{99.3} & 97.3 & 96.2 & 96.8 & 93.7 & 99.0 & 97.5 & 99.2 & 99.0 & 98.3 & 98.7 \\
leather & 90.0 & 98.6 & 98.0 & 99.4 & 99.2 & 97.4 & 99.2 & 97.6 & \textbf{99.5} & 95.5 & \textbf{99.5} & 99.7 & \textbf{99.5} & 99.3 \\
tile & 80.0 & 99.2 & 87.0 & 95.6 & 94.1 & 91.4 & 96.8 & 87.4 & 89.8 & 90.5 & \textbf{99.3} & 98.0 & 96.3 & 95.6 \\
wood & 81.0 & 96.4 & 93.0 & 95.3 & 94.9 & 90.8 & \textbf{99.6} & 88.5 & 95.8 & 95.0 & 90.7 & 96.7 & 97.0 & 95.0 \\ \hline
Average & 85.0 & 97.9 & 0.0 & 97.7 & 96.9 & 93.7 & 98.2 & 92.9 & 96.5 & 96.3 & 96.8 & \textbf{98.5} & 98.1 & 97.5 \\ \hline
bottle & 93.0 & \textbf{99.1} & 97.0 & 98.7 & 98.3 & 98.1 & 98.3 & 98.4 & 98.1 & 97.6 & 98.3 & 99.0 & 97.7 & 98.6 \\
cable & 94.0 & 95.5 & 92.0 & 97.4 & 96.7 & 96.8 & 97.5 & 97.2 & 93.2 & 90.0 & 96.0 & 97.6 & \textbf{98.4} & 98.4 \\
capsule & 90.0 & 94.3 & 99.0 & 98.7 & 98.5 & 95.8 & 98.6 & 99.0 & 98.6 & 97.4 & 97.6 & 99.0 & \textbf{99.1} & 98.8 \\
hazelnut & 84.0 & \textbf{99.7} & 99.0 & 98.9 & 98.2 & 97.5 & 98.7 & 99.1 & 98.9 & 97.3 & 97.6 & 98.9 & 99.1 & 98.7 \\
metal nut & 91.0 & \textbf{99.5} & 93.0 & 97.3 & 97.2 & 98.0 & 98.2 & 98.1 & 96.9 & 93.1 & 98.4 & 98.6 & 98.5 & 98.4 \\
pill & 93.0 & 97.6 & 97.0 & 98.2 & 95.7 & 95.1 & 97.3 & 96.5 & 96.5 & 95.7 & 98.5 & 99.0 & \textbf{99.2} & 97.4 \\
screw & 96.0 & 97.6 & 99.0 & 99.6 & 98.5 & 95.7 & 98.7 & 98.9 & 99.5 & 96.7 & 96.5 & 98.9 & \textbf{99.4} & \textbf{99.4} \\
toothbrush & 96.0 & 98.1 & 99.0 & 99.1 & 98.8 & 98.1 & 98.9 & 97.9 & 98.9 & 98.1 & 94.9 & \textbf{99.0} & 98.9 & 98.7 \\
transistor & \textbf{100.0} & 90.9 & 80.0 & 92.5 & 97.5 & 97.0 & 98.1 & 94.1 & 81.0 & 93.0 & 88.0 & 98.0 & 97.3 & 96.3 \\
zipper & 99.0 & 98.8 & 96.0 & 98.2 & 98.5 & 95.1 & 98.2 & 96.5 & 98.8 & \textbf{99.3} & 94.2 & 99.1 & 98.7 & 98.8 \\ \hline
Average & 93.6 & 97.1 & 0.0 & 97.9 & 97.8 & 96.7 & 98.3 & 97.6 & 96.0 & 95.8 & 96.0 & 98.7 & \textbf{98.6} & 98.4 \\ \hline
Average-all & 90.7 & 97.4 & 95.0 & 97.8 & 97.5 & 95.7 & 98.2 & 96.0 & 96.2 & 96.0 & 96.3 & \textbf{98.6} & 98.5 & 98.1 \\ \bottomrule
\end{tabular}
}
\caption{Anomaly detection performance (Pixel-level AUCROC) on MVTec AD dataset~\cite{bergmann2019mvtec}.}
\label{table:pixroc}
\end{table*}

\subsubsection{Inference Efficiency}
The experimental results of inference efficiency for SPADE, PatchCore, CFlow, CSFlow, DifferNet, FastFlow are shown in Table~\ref{table:complexity}. Since the CSFlow can not used for anomaly localization task, we report the result of anomaly detection for it, and others are evaluated under anomaly localization task. In the anomaly localization task, SPADE and PatchCore adopt a sliding window testing way to apply inference process on every patches, which slows down inference. DifferNet adopts a $64\times$ test time augmentation which leads to a slow inference speed. Benefit from the end-to-end inference, CFlow and FastFlow achieve faster inference. SPADE and PatchCore directly adopt the CNN backbone and do not need extra model parameters. Other methods use the flow model based on the CNN backbone. DifferNet introduces the most extra parameters and FastFlow achieves the fastest inference speed with minimal parameters introduced.

\begin{table}[tbp]
    \centering
    \begin{tabular}{c|ccccc}
        \hline
         Model & FPS &   Time (ms) & Extra parameters(MB)\\ 
         \hline
          SPADE & 0.67 & 1481 & 0 \\
          PatchCore & 5.88 & 159  & 0  \\
          CFlow & 14.9 & 56 & 81.3   \\
          CSFlow & 5.57 & 180 & 148.2  \\
          DifferNet & 2.53 & 395 & 172.1 \\
         FastFlow & \textbf{21.8} & \textbf{34} & 41.3 \\ 
         \bottomrule
       
    \end{tabular}
    \caption{Complexity comparison in terms of inference speed.}
    \label{table:complexity}
\end{table}




\subsubsection{Bad Cases with Ambiguity Label}
\begin{figure*}[t]
\centering
\includegraphics[width=0.9\linewidth]{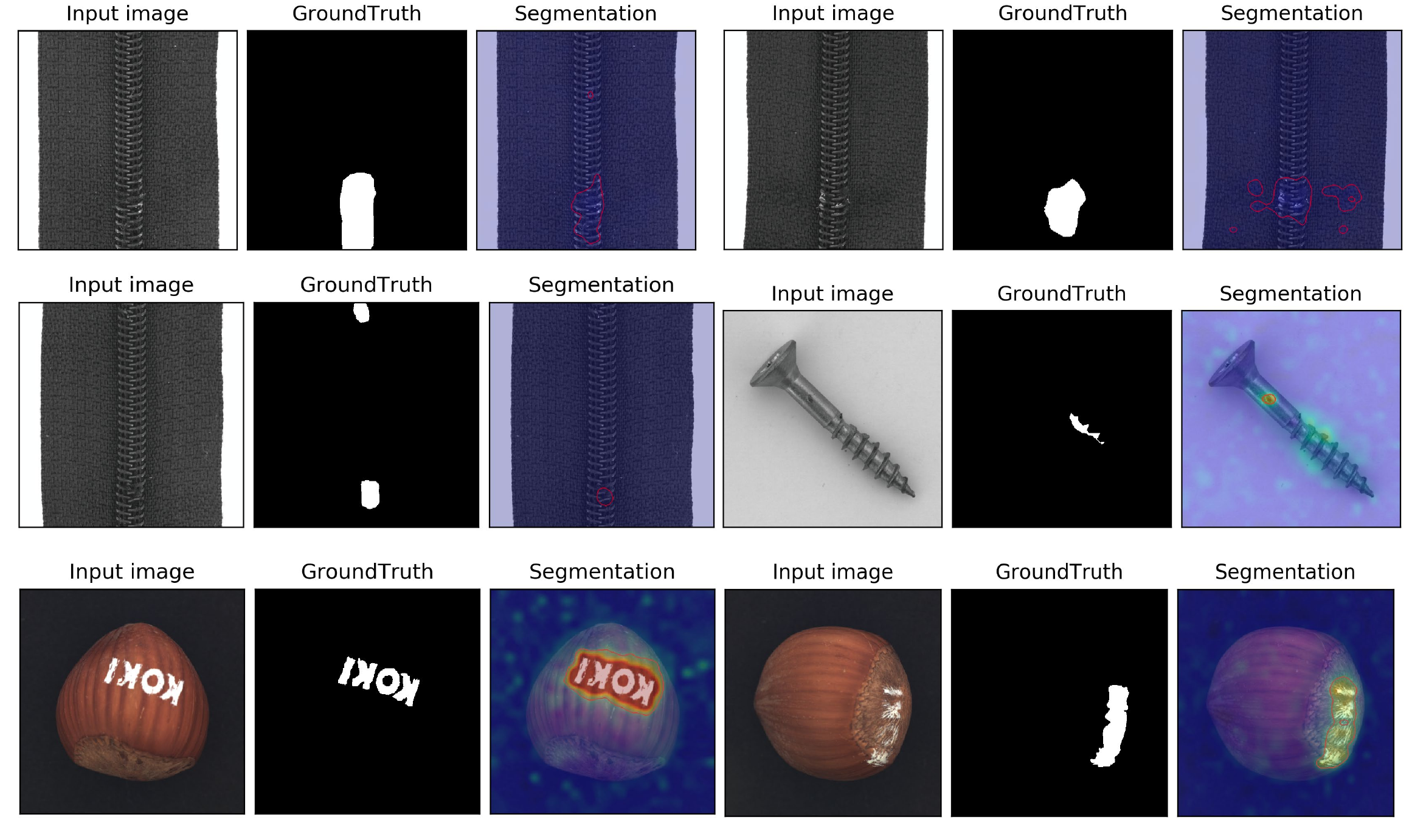} 
\caption{Bad cases caused by label ambiguity. In the first two rows, there are abnormal areas localized by our method while not labeled. In the last row of hazelnut, we show the label ambiguity of the ``print" subclass, in which one hazelnut print is labeled finely, while the other is labeled with a rough area.}
\label{fig:fig3}
\end{figure*}

We visualize bad cases cased by ambiguity label from FastFlow on MVTec AD dataset in Figure~\ref{fig:fig3}  some areas belong to abnormal but not be labeled, such as the ``scratch neck" for screw and the ``fabric interior" for zipper. Fixing these ambiguity annotations can further improve the performance.

\subsection{MVTec AD 3D}
We implement PaDim~\cite{defard2021Padim}, PatchCore~\cite{roth2021towards}, Fastflow~\cite{yu2021fastflow}, CFlow~\cite{gudovskiy2021cflow},DifferNet~\cite{rudolph2021same}, CSflow~\cite{rudolph2022fully} in 3D dataset with only using the RGB data. Except the DifferNet, we follow the best setting in these papers and the details of them shown in Table~\ref{tab:3d_imple}.

\begin{table}[htbp]
    \centering
    \begin{tabular}{c|c|c}
        \toprule
         Method &  Backbone & Input size \\
         \hline
         DifferNet & Alexnet &  [448, 224, 112] \\
         CSFlow & EfficientNetB5 & [768, 384, 192] \\
         PaDim & wide\_resnet50\_2  & 224 \\
         PatchCore& wide\_resnet50\_2 & 224 \\
         CFlow & wide\_resnet50\_2 & 224 \\
         FastFlow& wide\_resnet50\_2 & 224 \\
         \bottomrule
    \end{tabular}
    \caption{Implement details of the methods evaluted in MVTec 3D-AD datset.}
    \label{tab:3d_imple}
\end{table}

The performance comparison results of these methods are shown in Table~\ref{tab:3d_img} and~\ref{tab:3d_pix} in terms of image-level AUROC and pixel-level AUROC, respectively. The results in ``3D only" and ``3D+RGB" are the performances of all baseline models in~\cite{bergmann2021mvtec}.
We can learn that:
\begin{itemize}
    \item All state-of-the-art methods in 2D dataset can achieve significant improvement compared to all the baseline methods in~\cite{bergmann2021mvtec} when only RGB information is used, especially at the pixel-level.
    \item PatchCore and PaDim obtains the best performance in anomaly detection and anomaly localization, respectively. That indicates relatively straightforward methods achieve results.
    
\end{itemize}

\begin{table*}[tbp]
\begin{center}

\label{table:image-level}
\resizebox{\linewidth}{!}{

\begin{tabular}{l|llllll|llllll|llllll}
\hline
\multicolumn{1}{c|}{\multirow{3}{*}{Category}} & \multicolumn{6}{c|}{3d only}                                                                                                                               & \multicolumn{6}{c|}{3d + RGB}                                                                                                                              & \multicolumn{6}{c}{\multirow{2}{*}{RGB only}}                                                                                             \\ \cline{2-13}
\multicolumn{1}{c|}{}                          & \multicolumn{3}{c|}{voxel}                                                    & \multicolumn{3}{c|}{depth}                                                 & \multicolumn{3}{c|}{voxel}                                                    & \multicolumn{3}{c|}{depth}                                                 & \multicolumn{6}{c}{}                                                                                                                      \\ \cline{2-19} 
\multicolumn{1}{c|}{}                          & \multicolumn{1}{c}{GAN} & \multicolumn{1}{c}{AE} & \multicolumn{1}{c|}{VM}    & \multicolumn{1}{c}{GAN} & \multicolumn{1}{c}{AE} & \multicolumn{1}{c|}{VM} & \multicolumn{1}{c}{GAN} & \multicolumn{1}{c}{AE} & \multicolumn{1}{c|}{VM}    & \multicolumn{1}{c}{GAN} & \multicolumn{1}{c}{AE} & \multicolumn{1}{c|}{VM} & \multicolumn{1}{c}{PaDim} & \multicolumn{1}{c}{PatchCore} & \multicolumn{1}{c}{FastFlow} & \multicolumn{1}{c}{CFlow} & DifferNet & CSflow \\ \hline
bagel                                          & 0.383                   & 0.693                  & \multicolumn{1}{l|}{0.750} & 0.530                   & 0.468                  & 0.510                   & 0.680                   & 0.510                  & \multicolumn{1}{l|}{0.553} & 0.538                   & 0.648                  & 0.513                   & \textbf{0.975}                     & 0.912                         & 0.893                        & 0.880                     & 0.819     & 0.894  \\
cable gland                                    & 0.623                   & 0.425                  & \multicolumn{1}{l|}{0.747} & 0.376                   & 0.731                  & 0.542                   & 0.324                   & 0.54                   & \multicolumn{1}{l|}{0.772} & 0.372                   & 0.502                  & 0.551                   & 0.775                     & 0.902                         & 0.620                        & 0.858                     & 0.670     & \textbf{0.917}  \\
carrot                                         & 0.474                   & 0.515                  & \multicolumn{1}{l|}{0.613} & 0.607                   & 0.497                  & 0.469                   & 0.565                   & 0.384                  & \multicolumn{1}{l|}{0.484} & 0.580                   & 0.650                  & 0.477                   & 0.698                     & \textbf{0.885}                         & 0.795                        & 0.828                     & 0.612     & 0.749  \\
cookie                                         & 0.639                   & \textbf{0.790}                  & \multicolumn{1}{l|}{0.738} & 0.603                   & 0.673                  & 0.576                   & 0.399                   & 0.693                  & \multicolumn{1}{l|}{0.701} & 0.603                   & 0.488                  & 0.581                   & 0.582                     & 0.709                         & 0.426                        & 0.563                     & 0.484     & 0.668  \\
dowel                                          & 0.564                   & 0.494                  & \multicolumn{1}{l|}{0.823} & 0.497                   & 0.534                  & 0.609                   & 0.497                   & 0.446                  & \multicolumn{1}{l|}{0.751} & 0.430                   & 0.805                  & 0.617                   & 0.959                     & 0.952                         & 0.880                        & \textbf{0.986}                     & 0.634     & 0.938  \\
foam                                           & 0.409                   & 0.558                  & \multicolumn{1}{l|}{0.693} & 0.484                   & 0.417                  & 0.699                   & 0.482                   & 0.632                  & \multicolumn{1}{l|}{0.578} & 0.534                   & 0.522                  & 0.716                   & 0.663                     & 0.733                         & 0.728                        & 0.738                     & 0.689     & \textbf{0.897}  \\
peach                                          & 0.617                   & 0.537                  & \multicolumn{1}{l|}{0.679} & 0.595                   & 0.485                  & 0.450                   & 0.566                   & 0.550                  & \multicolumn{1}{l|}{0.480} & 0.642                   & 0.712                  & 0.450                   & \textbf{0.858}                     & 0.727                         & 0.651                        & 0.757                     & 0.655     & 0.603  \\
potato                                         & 0.427                   & 0.484                  & \multicolumn{1}{l|}{\textbf{0.652}} & 0.489                   & 0.549                  & 0.419                   & 0.579                   & 0.494                  & \multicolumn{1}{l|}{0.466} & 0.601                   & 0.529                  & 0.421                   & 0.535                     & 0.562                         & 0.560                        & 0.628                     & 0.600     & 0.419  \\
rope                                           & 0.663                   & 0.639                  & \multicolumn{1}{l|}{0.609} & 0.536                   & 0.564                  & 0.668                   & 0.601                   & 0.721                  & \multicolumn{1}{l|}{0.689} & 0.443                   & 0.540                  & 0.598                   & 0.832                     & 0.962                         & \textbf{0.982}                        & 0.970                     & 0.729     & 0.971  \\
tire                                           & 0.577                   & 0.583                  & \multicolumn{1}{l|}{0.690} & 0.521                   & 0.546                  & 0.520                   & 0.482                   & 0.413                  & \multicolumn{1}{l|}{0.611} & 0.577                   & 0.552                  & 0.623                   & 0.760                     & \textbf{0.768}                         & 0.613                        & 0.720                     & 0.536     & 0.726  \\ \cline{1-19}
\textbf{average}                                  & 0.538                   & 0.572                  & \multicolumn{1}{l|}{0.699} & 0.524                   & 0.546                  & 0.546                   & 0.517                   & 0.538                  & \multicolumn{1}{l|}{0.609} & 0.532                   & 0.595                  & 0.555                   & 0.764                     & \textbf{0.811}                         & 0.715                        & 0.793                     & 0.643     & 0.778  \\ \cline{1-19}
\end{tabular}

}
\end{center}
\caption{Anomaly detection performance (Image-level AUC) on MVTec AD 3D dataset~\cite{bergmann2021mvtec}}
\label{tab:3d_img}

\end{table*}

\begin{table*}[tbp]
\begin{center}

\label{table:image-level}
\resizebox{\linewidth}{!}{

\begin{tabular}{l|llllll|llllll|llll}
\hline
\multicolumn{1}{c|}{\multirow{3}{*}{Category}} & \multicolumn{6}{c|}{3D only}                                                                                                                               & \multicolumn{6}{c|}{3D + RGB}                                                                                                                              & \multicolumn{4}{c}{\multirow{2}{*}{RGB only}}                                                                        \\ \cline{2-13}
\multicolumn{1}{c|}{}                  & \multicolumn{3}{c|}{voxel}                                                    & \multicolumn{3}{c|}{depth}                                                 & \multicolumn{3}{c|}{voxel}                                                    & \multicolumn{3}{c|}{depth}                                                 & \multicolumn{4}{c}{}                                                                                                 \\ \cline{2-17} 
\multicolumn{1}{c|}{}                  & \multicolumn{1}{c}{GAN} & \multicolumn{1}{c}{AE} & \multicolumn{1}{c|}{VM}    & \multicolumn{1}{c}{GAN} & \multicolumn{1}{c}{AE} & \multicolumn{1}{c|}{VM} & \multicolumn{1}{c}{GAN} & \multicolumn{1}{c}{AE} & \multicolumn{1}{c|}{VM}    & \multicolumn{1}{c}{GAN} & \multicolumn{1}{c}{AE} & \multicolumn{1}{c|}{VM} & \multicolumn{1}{c}{PaDim} & \multicolumn{1}{c}{PatchCore} & \multicolumn{1}{c}{FastFlow} & \multicolumn{1}{c}{CFlow} \\ \hline
bagel                                  & 0.440                   & 0.260                  & \multicolumn{1}{l|}{0.453} & 0.111                   & 0.147                  & 0.280                   & 0.664                   & 0.467                  & \multicolumn{1}{l|}{0.510} & 0.421                   & 0.432                  & 0.388                   & \textbf{0.980}                     & 0.899                         & 0.880                        & 0.855                     \\
cable gland                            & 0.453                   & 0.341                  & \multicolumn{1}{l|}{0.343} & 0.072                   & 0.069                  & 0.374                   & 0.620                   & 0.750                  & \multicolumn{1}{l|}{0.331} & 0.422                   & 0.158                  & 0.321                   & 0.944                     & \textbf{0.953}                         & 0.752                        & 0.919                     \\
carrot                                 & 0.825                   & 0.581                  & \multicolumn{1}{l|}{0.521} & 0.212                   & 0.293                  & 0.243                   & 0.766                   & 0.808                  & \multicolumn{1}{l|}{0.413} & 0.778                   & 0.808                  & 0.194                   & 0.945                     & 0.957                         & 0.923                        & \textbf{0.958}                     \\
cookie                                 & 0.755                   & 0.351                  & \multicolumn{1}{l|}{0.697} & 0.174                   & 0.217                  & 0.526                   & 0.740                   & 0.550                  & \multicolumn{1}{l|}{0.715} & 0.696                   & 0.491                  & 0.570                   & \textbf{0.925}                     & 0.918                         & 0.812                        & 0.867                     \\
dowel                                  & 0.782                   & 0.502                  & \multicolumn{1}{l|}{0.680} & 0.160                   & 0.207                  & 0.485                   & 0.783                   & 0.765                  & \multicolumn{1}{l|}{0.680} & 0.494                   & 0.841                  & 0.408                   & 0.961                     & 0.930                         & 0.929                        &   \textbf{0.969}                     \\
foam                   & 0.378                   & 0.234  & \multicolumn{1}{l|}{0.284} & 0.128                   & 0.181                  & 0.314                   & 0.332                   & 0.473                  & \multicolumn{1}{l|}{0.279} & 0.252                   & 0.406                  & 0.282                   & \textbf{0.792}                     & 0.719                         & 0.646                        & 0.500                     \\
peach                                  & 0.392                   & 0.351                  & \multicolumn{1}{l|}{0.349} & 0.003                   & 0.164                  & 0.199                   & 0.582                   & 0.721                  & \multicolumn{1}{l|}{0.300} & 0.285                   & 0.262                  & 0.244                   & \textbf{0.966}                     & 0.920                         & 0.782                        & 0.889                     \\
potato                                 & 0.639                   & 0.658                  & \multicolumn{1}{l|}{0.634} & 0.042                   & 0.066                  & 0.388                   & 0.790                   & 0.918                  & \multicolumn{1}{l|}{0.507} & 0.362                   & 0.216                  & 0.349                   & \textbf{0.940}                     & 0.937                         & 0.615                        & 0.935                     \\
rope                                   & 0.775                   & 0.015                  & \multicolumn{1}{l|}{0.616} & 0.446                   & 0.545                  & 0.543                   & 0.633                   & 0.019                  & \multicolumn{1}{l|}{0.611} & 0.402                   & 0.716                  & 0.268                   & 0.937                     & \textbf{0.938}                         & 0.913                        & 0.904                     \\
tire                                   & 0.389                   & 0.185                  & \multicolumn{1}{l|}{0.346} & 0.075                   & 0.142                  & 0.385                   & 0.483                   & 0.170                  & \multicolumn{1}{l|}{0.366} & 0.631                   & 0.478                  & 0.331                   & 0.912                     & \textbf{0.929}                         & 0.550                        & 0.919                     \\ \hline
\textbf{average}                          & 0.583                   & 0.348                  & \multicolumn{1}{l|}{0.492} & 0.143                   & 0.203                  & 0.374                   & 0.639                   & 0.564                  & \multicolumn{1}{l|}{0.471} & 0.474                   & 0.481                  & 0.335                   & \textbf{0.930}                     & 0.910                         & 0.780                        & 0.871                     \\ \hline
\end{tabular}

}
\end{center}

\caption{Anomaly detection performance (Pixel-level PRO) on MVTec AD 3D dataset~\cite{bergmann2021mvtec}}
\label{tab:3d_pix}

\end{table*}

\section{Conclusion}
In this paper, we conduct a comprehensive comparison of existing mainstream unsupervised anomaly detection methods in MVTec AD dataset and MVTec AD-3D dataset under 2D and 3D tasks. In addition to giving the performance under the image-level and pixel-level metrics, the comparison of inference efficiency for some state-of-the-art methods are also provided. To explore which examples restrict existing methods from achieving all predictions correctly, we make a analysis of the label ambiguity in MVTec AD dataset. Moreover, this paper also conducts experiments using the existing state-of-the-art 2D methods on the new MVTec 3D-AD dataset, and we find existing method can also achieve a good performance in 3D dataset even only using RGB information.

{\small
	\bibliographystyle{ieee_fullname}
	\bibliography{egbib}
}

\end{document}